\documentclass[preprint,12pt]{elsarticle}

\usepackage{amssymb}
\usepackage{amsmath}
\usepackage{amsthm}

\usepackage{booktabs}
\usepackage{multirow}
\usepackage{array}
\usepackage{graphicx}
\usepackage{subcaption}
\usepackage{algorithm}
\usepackage{algpseudocode}
\usepackage{xcolor}
\usepackage{hyperref}
\usepackage{makecell}
\usepackage{booktabs}
\usepackage{multirow}
\usepackage{xcolor}
\usepackage{amssymb}
\usepackage{placeins}
\usepackage{subcaption}
\usepackage{threeparttable} 
\usepackage{geometry}
\journal{Communication in Transportation Research}

\begin{document}

\begin{frontmatter}



\title{SafeAlign-VLA: A Negative-Enhanced Safe Alignment Framework for Risk-Aware Autonomous Driving}

\author[1]{Kefei Tian\fnref{fn1}}
\ead{2251090@tongji.edu.cn}

\author[2]{Yuansheng Lian\fnref{fn1}}
\ead{lys22@mails.tsinghua.edu.cn}

\author[3]{Kai Yang}
\ead{kaiyang0401@gmail.com}

\author[4]{Xiangdong Chen}
\ead{cxd@nus.edu.sg}

\author[2]{Shen Li\corref{cor1}}
\ead{sli299@tsinghua.edu.cn}

\cortext[cor1]{Corresponding author.}
\fntext[fn1]{These authors contributed equally to this work.} 

\affiliation[1]{organization={College of Transportation, Tongji University},
            city={Shanghai},
            postcode={200092},
            country={China}}

\affiliation[2]{organization={Department of Civil Engineering, Tsinghua University},
            city={Beijing},
            postcode={100084},
            country={China}}

\affiliation[3]{organization={School of Vehicle and Mobility, Tsinghua University},
            city={Beijing},
            postcode={100084},
            country={China}}

\affiliation[4]{organization={Department of Civil and Environmental Engineering, National University of Singapore},
            city={Singapore},
            postcode={117576},
            country={Singapore}}
            
\begin{abstract}
End-to-end autonomous driving systems excel in common scenarios but struggle with safety-critical long-tail cases. Vision-Language-Action (VLA) models are promising due to their strong reasoning capabilities. However, most VLA-based approaches rely on positive expert demonstrations, rarely exploiting negative samples, leading to insufficient understanding of risky behaviors and safety boundaries.
To address this limitation, we propose SafeAlign-VLA, a unified negative-enhanced safe alignment framework that incorporates negative data into supervised learning and reinforcement learning. First, we develop a counterfactual safety pairing paradigm to generate structured safety labels and counterfactual positive trajectories from risky scenarios via counterfactual reasoning. Then, a two-stage training strategy is adopted: negative-enhanced supervised fine-tuning for failure feedback and trajectory correction, followed by anchor-based group relative policy optimization that uses positive and negative trajectories as contrastive anchors to steer sampling and penalize high-risk behaviors via group-relative advantages.
Experiments on NAVSIM and DeepAccident validate the proposed framework. SafeAlign-VLA achieves 89.1 PDMS on the NAVSIM v1 testset, improving over the baseline without negative data by 1.3\%. On DeepAccident, it reduces the collision rate to 3.36\%, while achieving 84.2\% language accuracy and 85.8\% risk prediction accuracy. These results demonstrate the effectiveness of the proposed negative-enhanced safe alignment framework for safe and robust autonomous driving.
\end{abstract}


\end{frontmatter}

\section{Introduction}
\label{sec:intro}

Autonomous driving aims to achieve safe and efficient transportation by integrating perception, prediction, decision-making, and control with minimal or no human intervention \cite{zhao2025survey, lian2026bap, li2026survey}. Traditional modular architectures, while offering strong interpretability, suffer from error accumulation and information loss across modules \cite{hussain2018autonomous, lian2026cdkformer}. End-to-end (E2E) approaches \cite{hu2022st, hu2023planning, chen2024vadv2} improve joint optimization and computational efficiency by directly mapping sensory inputs to control signals. However, most existing E2E autonomous driving methods primarily imitate human expert driving data or trajectory-level behaviors, while lacking logical reasoning behind driving behaviors. \cite{chen2024end, codevilla2019exploring, omeiza2021explanations, hu2026lift}.

To address these limitations, recent work introduces Vision-Language-Action (VLA) models by integrating Vision-Language Models (VLMs) into end-to-end driving frameworks \cite{hwang2024emma, sima2024drivelm, marcu2024lingoqa}. By grounding pixel-level observations in high-level semantic representations, VLA models enable structured reasoning such as chain-of-thought (CoT) reasoning \cite{wang2025multimodal, nie2024reason2drive}, substantially improving generalization over prior E2E methods. Building on large-scale multimodal pretraining, these models leverage commonsense knowledge and linguistic reasoning to handle unfamiliar or long-tail scenarios more robustly \cite{xu2024drivegpt4, shao2024lmdrive, zhou2026opendrivevla}. Recent advances further extend VLA frameworks with counterfactual reasoning \cite{tian2024drivevlm, wang2025omnidrive} and preference-based policy alignment \cite{shang2025drivedpo}, demonstrating promising safety improvements in challenging conditions.

Despite these advances, existing VLA-based driving systems still rely predominantly on imitation learning from positive expert demonstrations, which limits robustness and generalization in unseen or safety-critical long-tail scenarios. Real-world negative data, such as collisions and near-misses, contain valuable signals about unsafe behaviors to avoid and recovery strategies under challenging conditions, yet they are largely overlooked in current training pipelines \cite{jaeger2023hidden, lu2023imitation, chen2024end}. Without exposure to negative cases, models often fail to learn clear safety boundaries, leading to unreliable decision-making and degraded robustness in high-risk situations \cite{wang2025enhancing, patrikar2025case}.

To address these limitations, we propose SafeAlign-VLA, a unified negative-enhanced safe alignment framework that explicitly integrates negative signals into both supervision and policy optimization, enabling safer and more robust decision-making. Our key contributions are threefold:

\begin{enumerate}
    \item We propose a two-stage safe alignment framework that leverages negative samples across both supervised fine-tuning (SFT) and reinforcement learning (RL) to enable negative analysis, trajectory correction, and risk-aware decision-making. To support this framework, we design counterfactual safety pairing, a data construction paradigm that extracts structured safety-critical supervision information and generates positive alternatives from high-risk scenarios.
   
    \item We introduce an anchor-based group relative policy optimization (GRPO) method that encourages the policy to approach positive behaviors while avoiding negative modes. It leverages positive and negative trajectories as anchors, together with a feedback-driven refinement strategy, to improve sample quality for structured reward shaping and effective policy optimization.

    \item Extensive experiments on the NAVSIM and DeepAccident benchmarks validate the proposed framework. Evaluation on the NAVSIM v1 testset shows that SafeAlign-VLA achieves SOTA performance. On DeepAccident, the method demonstrates superior risk recognition and proactive safety decision-making ability under long-tail driving scenarios.
\end{enumerate}

\section{Related Work}
\label{sec:related}

\subsection{VLA Models for Autonomous Driving}
\label{subsec:vlm}
To address the limited generalization of traditional modular driving systems in complex scenarios, recent studies have proposed VLA models, which unify perception, semantic reasoning, and control into a single policy \cite{sapkota2025vision}. Using pre-training foundation models on large-scale vision and language data, VLA enables direct mapping from multimodal input to driving actions, achieving strong generalization across benchmarks \cite{hu2025vision}.

Existing VLA approaches can be broadly categorized into two types. The first focuses on joint representation learning, where perception and planning are optimized through shared features \cite{jiang2024senna, jiang2025alphadrive, marcu2024lingoqa, tian2024drivevlm}. For example, DriveVLM \cite{tian2024drivevlm} integrates visual understanding and trajectory planning within a unified VLM framework, leveraging shared scene representations to enable coherent perception-to-planning reasoning. The second type emphasizes end-to-end policy learning, directly mapping multimodal inputs to executable actions or trajectories \cite{fu2025orion, zhao2025sce2drivex, li2025recogdrive, hwang2024emma}. OpenDriveVLA \cite{zhou2026opendrivevla} enhances the spatial reasoning of end-to-end driving by introducing hierarchical 3D visual representations and ego-environment interaction modeling. SimLingo \cite{renz2025simlingo} achieves breakthrough performance in vision-only closed-loop tasks by reinforcing the semantic alignment between linguistic commands and the vehicle's action space.

To improve decision reliability, recent work incorporates CoT reasoning, introducing structured pipelines to guide decision-making \cite{sima2024drivelm}. Furthermore, counterfactual reasoning has been explored to improve safety by analyzing potential outcomes of alternative actions \cite{wang2025omnidrive, ccot}. 

\subsection{Learning from Negative Samples for Risk-Aware Driving}
\label{subsec:negative}
In risk-aware autonomous driving, most existing approaches rely on expert demonstrations or safe trajectories as supervision signals, training models via behavior cloning or imitation learning to approximate human driving behavior \cite{pan2017agile, cheng2024pluto, sun2025sparsedrive, wu2022trajectory}. However, these methods are dominated by positive samples and lack coverage of high-risk and negative modes, leading to degraded performance in out-of-distribution or complex scenarios \cite{jaeger2023hidden, chen2024end}. In contrast, negative samples, including collision events and near-miss data, provide critical safety-related information essential for defining safety boundaries and enhancing model robustness \cite{patrikar2025case,lu2025controllable}.

To alleviate this limitation, recent studies have explored incorporating negative samples into training. Some works introduce synthetic or real-world collision data into the training pipeline \cite{bansal2018chauffeurnet, wang2025enhancing}, while others leverage counterfactual reasoning over accident cases to evaluate candidate actions \cite{patrikar2025case}. In addition, preference-based methods such as DriveDPO \cite{shang2025drivedpo} construct safe-unsafe trajectory pairs and perform contrastive optimization to align policies with safer behaviors. SpanVLA \cite{zhou2026spanvla} leverages positive, negative, and recovery behaviors through a GRPO-based post-training method with real-world takeover data, improving policy robustness and recovery capacity in challenging driving scenarios.

Despite these advances, two key limitations remain. First, negative cases are often treated as coarse labels without fine-grained causal analysis, limiting the model’s ability to reason about the underlying causes of failure and generate corrective actions \cite{cheng2025adreft}. Second, although recent RL methods leverage positive and negative samples \cite{zhou2026spanvla}, they still lack a unified pipeline for automated positive-negative pair generation and self-feedback optimization, limiting iterative safety-driven self-improvement. 

To address these issues, we propose SafeAlign-VLA, a two-stage unified negative-enhanced safe alignment framework. The supervised stage incorporates negative samples for diagnosis and trajectory correction, while the RL stage introduces an anchor-based contrastive GRPO method, using positive and negative trajectories as corresponding anchors to guide policy optimization toward safer behaviors, together with feedback-based sample refinement to improve exploration quality and training effectiveness.

\section{Methodology}
\label{sec:method}

\subsection{Problem Formulation}
\label{subsec:formulation}

In this paper, we formulate the safe planning task as a sequential decision-making problem. Given the multimodal input, which comprises visual observations $\mathbf{V}$ and language prompts $\mathbf{L}$, the goal of our model is to learn a policy $\pi_\theta(\mathbf{T}, \mathbf{A} | \mathbf{V}, \mathbf{L})$ parameterized by $\theta$, which maps the input to a dual output space including both semantic reasoning text $\mathbf{T}$ and action $\mathbf{A}$.

Crucially, to ensure safety in long-tail and high-risk scenarios, the policy $\pi_\theta$ is optimized not only to imitate expert positive demonstrations ($\tau_{\text{pos}}$) but also to explicitly recognize and distance itself from high-risk negative trajectories ($\tau_{\text{neg}}$).


\subsection{Framework Overview}
\label{subsec:overview}

\begin{figure}[t!]
    \centering
    \includegraphics[width=\textwidth]{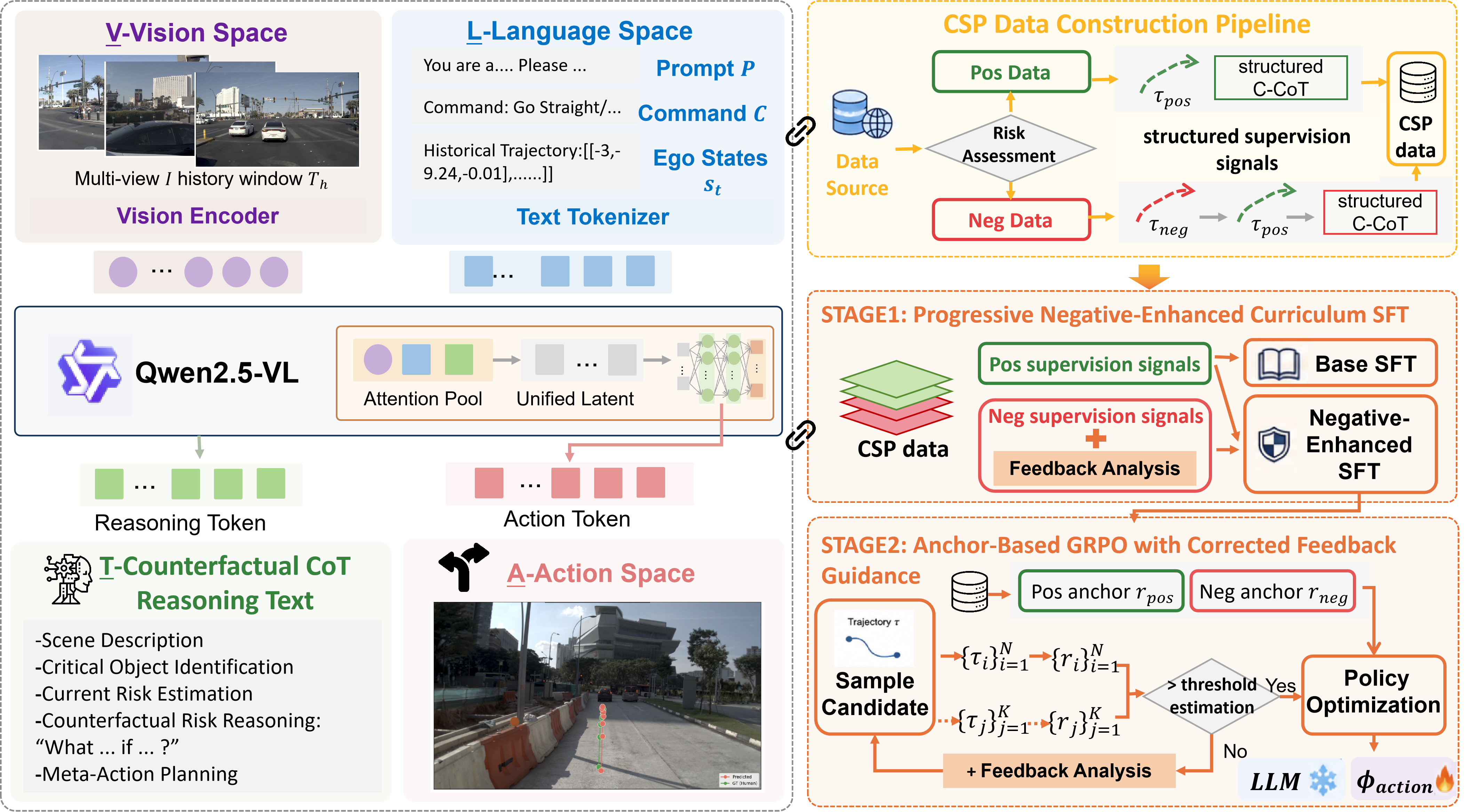} 
    \caption{The proposed SafeAlign-VLA framework for risk-aware autonomous driving. The framework consists of a reasoning-driven VLA planning architecture, a counterfactual safety pairing module for constructing positive-negative supervision data, and a two-stage negative-enhanced safe alignment strategy that combines supervised fine-tuning and reinforcement learning.}
    \label{fig:overall_framework}
\end{figure}

This study proposes SafeAlign-VLA, a unified negative-enhanced safe alignment framework for autonomous driving. 
As illustrated in Figure~\ref{fig:overall_framework}, the framework consists of three tightly coupled components: a reasoning-driven VLA planning architecture, a counterfactual safety pairing data construction paradigm, and a two-stage post-training strategy, enabling the model to learn not only how to imitate expert behaviors but also how to avoid unsafe decisions.

At each time step $t$, the visual input $\mathbf{V}$ contains three-view camera images that capture the surrounding environment within a past horizon $T_h$. The language input $\mathbf{L}$ is defined as a comprehensive text prompt $\mathbf{L}=\{\mathbf{s_t},\mathbf{P},\mathbf{C}\}$, where $\mathbf{s}_t$ represents the ego historical trajectory states within a past horizon $T_h$, $\mathbf{P}$ is the task-specific prompt, and $\mathbf{C}$ denotes the high-level driving command.
To enhance interpretability and safety awareness, the task-specific prompt $\mathbf{P}$ incorporates a C-CoT reasoning process \cite{ccot} with five sequential steps: scene description, critical object identification, current risk estimation, counterfactual risk reasoning, and meta-action planning. The meta-action planning step produces two consecutive high-level actions as textual outputs for short-term ($t$–$T_s$) and long-term ($T_s$–$T_p$) planning, respectively. Each meta-action consists of a speed decision and a directional decision, where the speed decision is selected from \{maintain speed, accelerate, decelerate\}, and the directional decision is selected from \{ keep lane, turn left, turn right, change lane towards left, change lane towards right\}.

As shown in the VLA planning branch on the left side of Fig.~\ref{fig:overall_framework}, the visual input $\mathbf{V}$ in the V-Space is encoded by the vision encoder, while the language input $\mathbf{L}$ in the L-Space is processed by the text tokenizer. These representations are then fed into the LLM backbone for multimodal feature fusion. The LLM backbone sequentially generates the semantic reasoning sequence $\mathbf{T}$ adhering to the C-CoT schema. 
For trajectory generation, an attention pooling module $\phi_{\text{attn\_pool}}$ is then applied to the final-layer hidden states $\mathbf{H}^{L}$ of the LLM backbone to obtain a unified latent feature. The action head $\phi_{\text{action}}$ then regresses future waypoints $\mathbf{A}=(x_t,y_t)_{t=1}^{T_p}$ from the feature, enabling end-to-end transformation from multimodal perception and reasoning representation to physical driving actions.

To provide structured supervision beyond standard expert demonstrations, we further design a counterfactual safety pairing data construction paradigm as the core data engine (Fig.~\ref{fig:overall_framework}, top right). The paradigm automatically identifies high-risk samples and synthesizes paired positive-negative supervision signals through counterfactual replanning.

Finally, SafeAlign-VLA is optimized using a two-stage learning paradigm (Fig.~\ref{fig:overall_framework}, bottom right). Specifically, stage 1 implements a progressive curriculum SFT that equips the model with foundational scene understanding ability before introducing structured negative supervision for trajectory-level self-correction. Stage 2 introduces an anchor-based GRPO to finetune the action head. A contrastive anchor pair is introduced to optimize relatively safe rewards and suppress unsafe modes, and a corrected feedback guidance is developed to iteratively refine sample quality for effective policy exploration.
Through the synergy of architecture, data, and optimization, SafeAlign-VLA produces safe, interpretable, and robust driving behaviors in complex traffic environments.

\subsection{Counterfactual Safety Pairing}
\label{subsec:generation}

\begin{algorithm}[t!]
\caption{CSP Data Construction Paradigm}
\label{alg:csp_pipeline}
\begin{algorithmic}[1]
\renewcommand{\algorithmicrequire}{\textbf{Input:}}
\renewcommand{\algorithmicensure}{\textbf{Output:}}
\Require Raw driving dataset $\mathcal{D}_{\text{raw}}$, prediction horizon $T_p$, safety threshold $TTC_{\text{risk}}$, number of candidates $K$, planning-quality metric $S(\cdot)$
\Ensure Risk-aware aligned dataset $\mathcal{D}_{\text{CSP}}$

\State Initialize $\mathcal{D}_{\text{CSP}} \leftarrow \emptyset$
\For{each sequence in $\mathcal{D}_{\text{raw}}$}
    \For{each time step $t$}
        \State Extract observation $o_t$ and observed trajectory $\tau_{\text{obs}}$ over $(t, t+T_p)$
        \State Compute $\textit{TTC}_{\min}(t) = \min_{t' \in (t,\, t+T_p)} \textit{TTC}(t')$
        
        \Statex \vspace{-0.5em} \State \Comment{\textit{Step 1: Safety Labeling}}
        \If{$\textit{TTC}_{\min}(t) \geq TTC_{\text{risk}}$}
            \State $y_t \leftarrow \text{Pos}$
        \Else
            \State $y_t \leftarrow \text{Neg}$
        \EndIf
        
        \Statex \vspace{-0.5em} \State \Comment{\textit{Step 2: Trajectory Supervision Assignment}}
        \If{$y_t = \text{Pos}$}
            \State $\tau_{\text{pos}} \leftarrow \tau_{\text{obs}}$, \quad $\tau_{\text{neg}} \leftarrow \emptyset$
        \ElsIf{$y_t = \text{Neg}$}
            \State $\tau_{\text{neg}} \leftarrow \tau_{\text{obs}}$ \Comment{\textit{Record unsafe execution}}
            \State Sample candidates $\{\tilde{\tau}_k\}_{k=1}^{K} \sim p(\tau \mid o_t)$ via longitudinal replanning
            \State $\tau_{\text{pos}} \leftarrow \arg\max_{\tilde{\tau}_k \in \{\tilde{\tau}_k\}} S(\tilde{\tau}_k)$ \Comment{\textit{Optimal counterfactual}}
        \EndIf

        \Statex \vspace{-0.5em} \State \Comment{\textit{Step 3: Semantic Reasoning Construction (C-CoT)}}
        \State Generate structured C-CoT JSON $C_{\text{CoT}}$ including:
        \State \quad $\bullet$ Scene description 
        \State \quad $\bullet$ Critical object identification
        \State \quad $\bullet$ Risk estimation ($y_t$) 
        \State \quad $\bullet$ Counterfactual reasoning
        \State \quad $\bullet$ Meta-action planning conditioned on $\tau_{\text{pos}}$
        
        \Statex \vspace{-0.5em} \State \Comment{\textit{Step 4: Data Aggregation}}
        \State $\mathcal{D}_{\text{CSP}} \leftarrow \mathcal{D}_{\text{CSP}} \cup \{(o_t, C_{\text{CoT}}, \tau_{\text{pos}}, \tau_{\text{neg}}, y_t)\}$
    \EndFor
\EndFor
\State \Return $\mathcal{D}_{\text{CSP}}$
\end{algorithmic}
\end{algorithm}

Currently, most existing open-source datasets predominantly provide expert demonstrations, but lack sufficient coverage of failure modes, which might limit the model's ability to identify safety boundaries.
However, we argue that autonomous driving VLA models benefit not only from imitating expert demonstrations (positive samples) but also from explicitly learning to avoid high-risk behaviors (negative samples). Within this framework, negative samples are broadly defined as behaviors that violate safety constraints, encompassing both real-world collision data and synthesized high-risk trajectories.
In this study, we propose counterfactual safety pairing (CSP), a unified data construction paradigm designed to derive structured supervision signals from large-scale driving datasets (Algorithm~\ref{alg:csp_pipeline}). Specifically, CSP identifies failure modes based on safety measures and generates counterfactual safe alternatives to form a positive-negative data pair.

For each time step $t$, we evaluate the minimum time-to-collision between the ego vehicle and surrounding agents within the future interval $(t, t+T_p)$, denoted as $TTC_{\min}(t)$. Based on this metric, we define a binary safety label using a predefined TTC threshold $TTC_{\text{risk}}$:
\begin{equation}
y_t =
\begin{cases}
\text{Pos}, & \text{if } \;TTC_{\min}(t) \geq TTC_{\text{risk}},  \\
\text{Neg}, & \text{if } \;TTC_{\min}(t) < TTC_{\text{risk}}.
\end{cases}
\end{equation}

This labeling criterion provides a consistent and interpretable definition of interaction safety, which serves as the basis for constructing downstream supervision signals across modalities.
For trajectory supervision, positive and negative scenarios are treated asymmetrically. In positive cases, the ground-truth future trajectory is used directly as the supervision target $\tau_{\text{pos}}$. In negative cases, the observed trajectory $\tau_{\text{neg}}$ represents unsafe execution. And a counterfactual positive trajectory is generated by replanning feasible motion along the ground-truth lane centerline while preserving route intent, which serves as $\tau_{\text{pos}}$. Longitudinal acceleration is scaled proportionally to the current speed to ensure physical plausibility. Candidate trajectories are constructed via piecewise longitudinal control over two temporal stages, producing multiple feasible motion alternatives under the same observation $o_t$:
\begin{equation}
\{\tilde{\tau}_k\}_{k=1}^{K} \sim p(\tau \mid o_t)
\end{equation}

Each candidate is then evaluated by the planning-quality score $S(\cdot)$, which is defined as the PDM score (PDMS)\cite{dauner2024navsim}.  Specifically, given a predicted trajectory, a forward simulation is performed and multiple rule-based indicators are calculated, including no at-fault collision (NC), drivable area compliance (DAC), ego progress (EP), time-to-collision (TTC) and comfort (C), which are aggregated as $PDMS = \text{NC} \times \text{DAC} \times (5 \cdot \text{EP} + 5 \cdot \text{TTC} + 2 \cdot \text{C}) / 12$. The optimal counterfactual positive trajectory is selected as $\tau_{\text{pos}}$:
\begin{equation}
\tau_{\text{pos}} = \arg\max_{\tilde{\tau}_k \in \{\tilde{\tau}_k\}_{k=1}^{K}} S(\tilde{\tau}_k).
\end{equation}

For language-based reasoning supervision, we construct structured C-CoT annotations in JSON format, which encode step-wise reasoning including scene description, critical object identification, current risk estimation based on $y_t$, counterfactual risk reasoning, and meta-action generation conditioned on $\tau_{\text{pos}}$, thereby enabling interpretable and safety-aligned supervision for the language head.

As a result, each negative case yields a paired supervision signal $(\tau_{\text{neg}}, \tau_{\text{pos}})$, enabling trajectory-level learning under both conditions and providing explicit contrastive supervision to distinguish desirable behaviors from negative-inducing actions.

\subsection{Progressive Negative-Enhanced Curriculum SFT}
\label{subsec:Negative-Enhanced SFT}
\begin{figure*}[t]
    \centering
    \includegraphics[width=\textwidth]{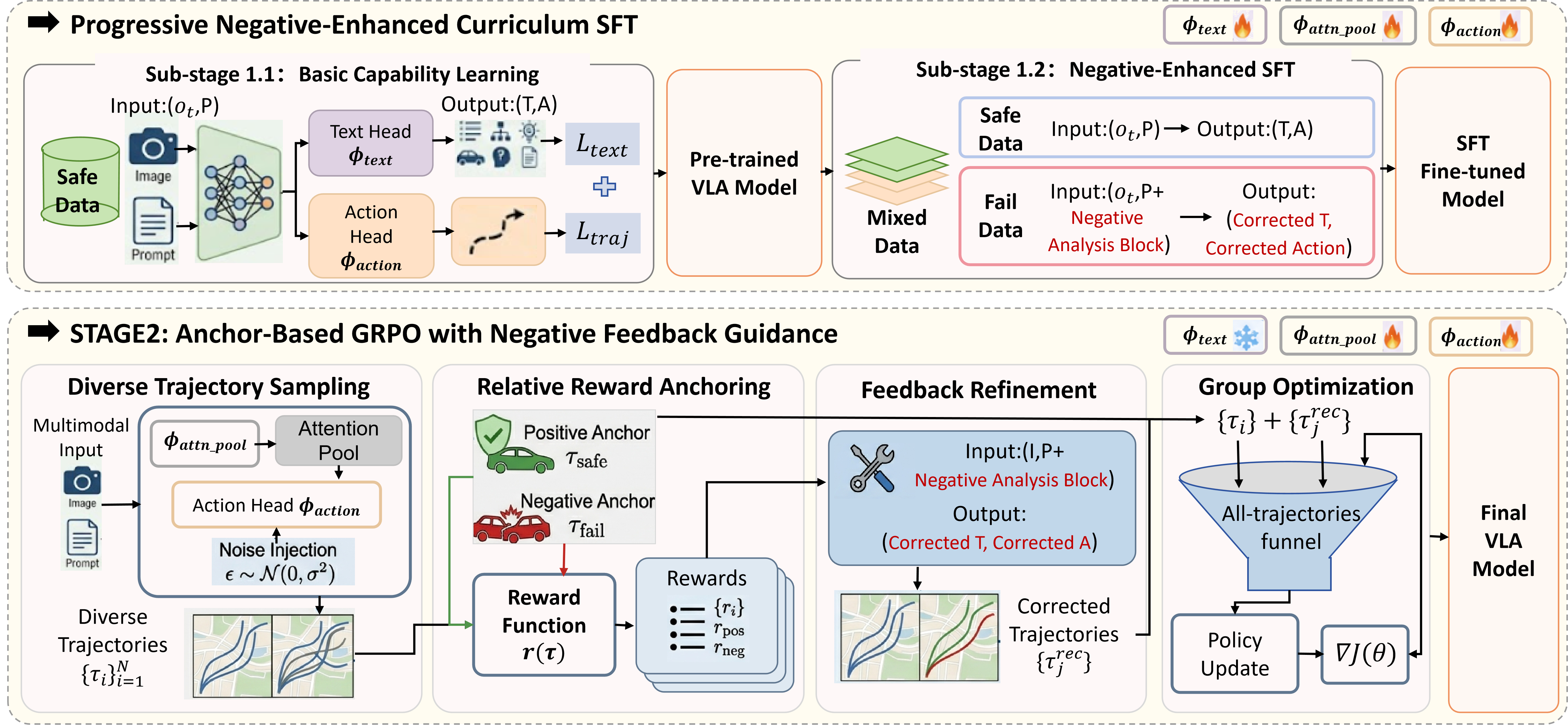}
    \caption{Overall framework of the proposed method. 
    Stage 1 performs negative-enhanced curriculum SFT, including basic capability learning and negative-enhanced mixed SFT. 
    Stage 2 adopts RL with feedback refinement using anchor-based GRPO, incorporating diverse trajectory sampling, reward anchoring, feedback refinement, and group optimization.}
    \label{fig:framework}
\end{figure*}

We formulate the SFT process as a curriculum learning paradigm \cite{bengio2009curriculum}, where the model first learns from positive demonstrations and is then progressively exposed to negative cases with corrective supervision, as illustrated in the upper part of Figure~\ref{fig:framework}. Accordingly, the first stage focuses on acquiring fundamental scene-understanding and trajectory-planning capabilities, while the second stage emphasizes learning to correct trajectories through feedback analysis.

In the first stage, the model is trained on standard positive data using a multi-task joint loss function:

\begin{equation}
L_\text{total} = L_\text{text} + \alpha \cdot L_\text{traj}, \quad L_\text{traj} = \frac{1}{T_p} \sum_{t=1}^{T_p} \left\| \mathbf{A}_t - \hat{\mathbf{A}}_t \right\|^2
\end{equation}
where $L_\text{text}$ denotes the supervised loss for text generation, ensuring the model produces outputs $\mathbf{T}$ consistent with reference answers under multimodal input. $L_\text{traj}$ represents the mean squared error (MSE) loss for trajectory prediction, supervising the action head outputs $\mathbf{A}$. A gradient scaling mechanism is applied to balance the gradient magnitudes, with $\alpha$ adjusting the trajectory loss to remain on a similar scale as the text loss, ensuring stable multi-task optimization.

\begin{figure}[t]
    \centering
    \includegraphics[width=\linewidth]{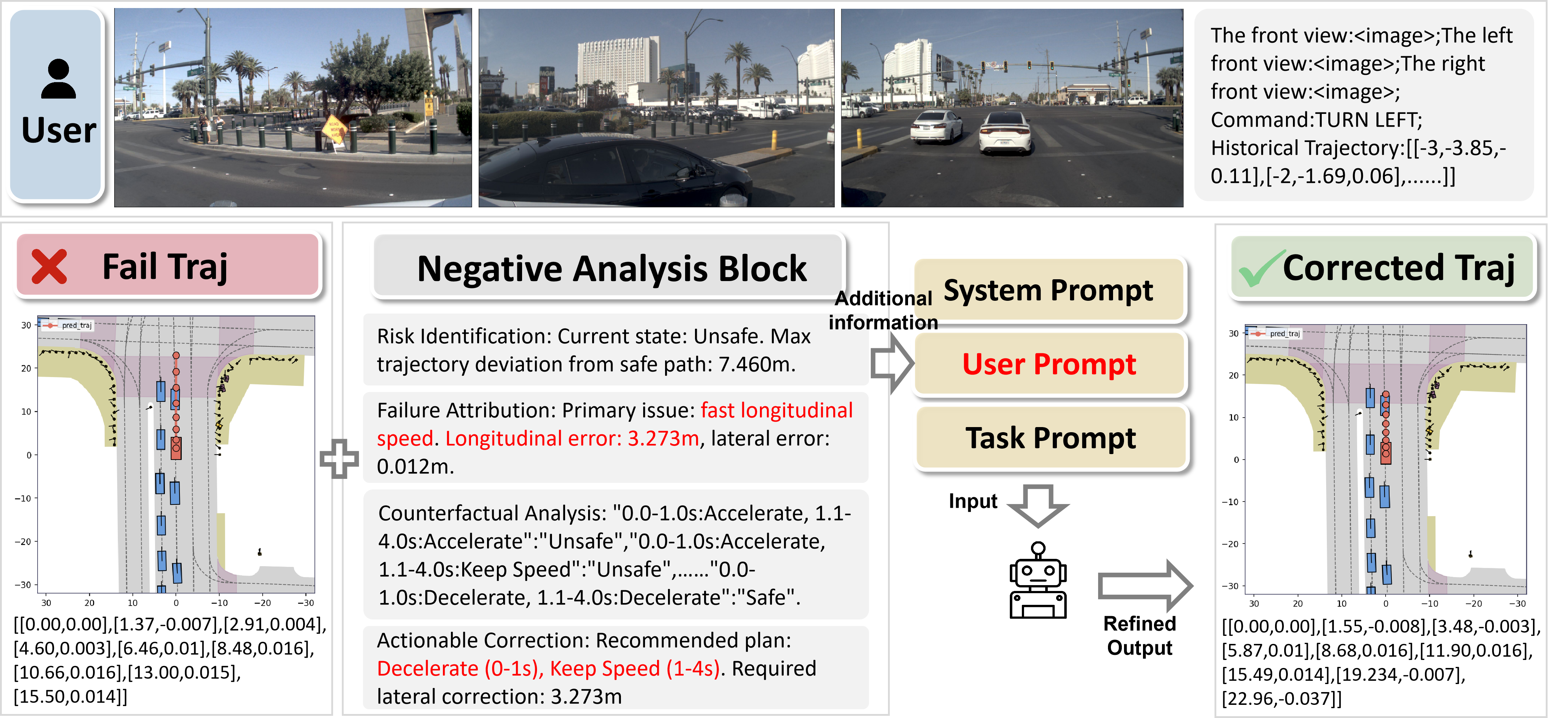}
    \caption{
        Overview of the negative-enhanced mixed SFT stage. Given input data and the paired supervision signal $(\tau_{\text{neg}}, \tau_{\text{pos}})$, the framework constructs a structured negative analysis block comprising four levels of semantic guidance: risk identification, failure attribution, counterfactual analysis, and actionable correction. This block and the negative trajectory $\tau_{\text{neg}}$ are then injected into the user prompt alongside the system and task prompts. The model is trained to produce a corrected positive trajectory (right) from the originally negative one (left).
    }
    \label{fig:negative_aware_sft}
\end{figure}

The second stage, negative-enhanced SFT, further enables the model to recognize risks and correct actions based on a feedback prompt. Training data consists of an equal mix of normal positive scenarios and negative correction scenarios. Normal scenarios ensure accurate scene understanding and positive trajectory outputs, while negative scenarios introduce the negative trajectory($\tau_\text{neg}$) and a structured negative analysis block in the prompt. The block provides four levels of semantic guidance: risk identification, failure attribution, counterfactual analysis, and actionable correction. As illustrated in Figure~\ref{fig:negative_aware_sft}, the model is supervised to use the original multimodal input, together with \(\tau_{\text{neg}}\) and the negative analysis, to correct negative behaviors and output the corresponding safe trajectory.

Specifically, risk identification declares the current scene as unsafe and quantifies the deviation of the trajectory from the safe trajectory; failure attribution performs a fine-grained decomposition of trajectory errors into longitudinal and lateral components to identify the primary cause; counterfactual analysis establishes explicit associations between alternative actions and their resulting outcomes, capturing the causal relationship between decisions and consequences; and actionable correction generates time-aligned control adjustments based on $\tau_{\text{pos}}$, with explicit correction magnitudes aligned with the target trajectory. This allows the model to understand the causal mechanisms behind errors and perform trajectory corrections, producing safe and executable trajectories.  

Together, the curriculum SFT allows the model to first establish foundational capabilities for scene understanding and trajectory planning, and then develop the ability to correct trajectories based on supplementary analysis, providing a solid basis for closed-loop feedback in subsequent RL.

\subsection{Anchor-Based GRPO with Failure Feedback Guidance}

To further strengthen the model’s robustness and proactive safety decision-making in complex risk scenarios, we introduce an RL stage guided by failure feedback, as illustrated in the lower part of Figure~\ref{fig:framework}. This stage extends the capabilities of the model beyond those acquired in SFT, enabling active exploration of the decision space and learning to perform safe behaviors.
To enhance the mapping from multimodal inputs to actions $\mathbf{A}$, we freeze the ViT and LLM backbone and fine-tune only the attention pooling layer $\phi_\text{attn\_pool}$ and action head $\phi_\text{action}$. This preserves high-level semantic reasoning while optimizing low-level action generation.

\subsubsection{Anchor-Based GRPO Training Pipeline}

\begin{algorithm}[t!]
\caption{Anchor-Based GRPO with Failure Feedback Guidance}
\label{alg:rl_pipeline}
\begin{algorithmic}[1]
\renewcommand{\algorithmicrequire}{\textbf{Input:}}
\renewcommand{\algorithmicensure}{\textbf{Output:}}
\Require CSP dataset $\mathcal{D}_{\text{CSP}}$, reward $r(\cdot)$, candidate count $N$, feedback size $K$, threshold $\delta$
\Ensure Updated policy $\theta = \{\phi_\text{pool},\ \phi_\text{action}\}$

\For{each scene in $\mathcal{D}_{\text{CSP}}$}
    \State Extract $o_t,\ \tau_{\text{pos}},\ \tau_{\text{neg}}$ from scene
    \State Obtain visual $I$ and command $P$ from $o_t$; Encode reasoning $T = \text{LLM}(I, P)$
    
    \Statex \vspace{-0.5em} \State \Comment{\textit{Step 1: Diverse Trajectory Sampling}}
    \For{$i = 1$ to $N$}
        \State Sample $\epsilon_i \sim \mathcal{N}(0, \sigma^2)$; \quad $\mathbf{f}_\text{base} = \phi_\text{attn\_pool}(I, P, T)$; \quad
        $\mathbf{f}_\text{i} = \mathbf{f}_\text{base} +\xi_\text{i}$
        \State Generate $\tau_i = \phi_\text{action}(\mathbf{f}_\text{i})$; \quad Evaluate reward $r_i = r(\tau_i)$
    \EndFor

    \Statex \vspace{-0.5em} \State \Comment{\textit{Step 2: Anchor Reward Evaluation}}
    \State Compute $r_{\text{pos}} = r(\tau_{\text{pos}})$, $\ r_{\text{neg}} = r(\tau_{\text{neg}})$

    \Statex \vspace{-0.5em} \State \Comment{\textit{Step 3: Feedback Refinement}}
    \State $\mathcal{T}_{\text{fb}} \leftarrow \emptyset$,\ $R_{\text{fb}} \leftarrow \emptyset$
    \If{$\max \{ r_i \}_{i=1}^N < \delta$}
        \State Analyze worst sample $\tau_{\arg\min_i r_i}$ vs. $\tau_{\text{pos}}$
        \State Sample corrected $\{\tau_j^{\text{rec}}\}$ where $r_j^{\text{rec}} > \max_i r_i$; truncate to $K$
        \State $\mathcal{T}_{\text{fb}} \leftarrow \{\tau_j^{\text{rec}}\}_{j=1}^{K}$,\quad $R_{\text{fb}} \leftarrow \{r_j^{\text{rec}}\}_{j=1}^{K}$
    \EndIf
    
    \Statex \vspace{-0.5em} \State \Comment{\textit{Step 4: Advantage Normalization}}
    \State Compute $\mu, \sigma$ from $R_{\text{samples}} = \{r_i\} \cup R_{\text{fb}}$ \hfill\Comment{\textit{Anchors excluded from $\mu, \sigma$}}
    \State $\hat{A}_k = ({r_k - \mu})/{\sigma}$,\quad $\forall\, r_k \in R_{\text{samples}} \cup \{r_{\text{pos}},\ r_{\text{neg}}\}$

    \Statex \vspace{-0.5em} \State \Comment{\textit{Step 5: Policy Optimization}}
    \State Update $\theta$ by minimizing contrastive loss $\mathcal{L}$:
    \State \small $ \mathcal{L} = \frac{1}{|\mathcal{T}|} \sum_{\tau_k \in \mathcal{T}}
       \begin{cases}
            \hat{A}_k \cdot \mathcal{L}_{\mathrm{Huber}}(\tau_k, \tau_{\mathrm{pos}}),
            & \hat{A}_k \ge 0, \\
            |\hat{A}_k| \cdot \max(0, m - d(\tau_k,\tau_{\mathrm{neg}})),
            & \hat{A}_k < 0.
        \end{cases} $
\EndFor
\State \Return $\theta$
\end{algorithmic}
\end{algorithm}

Our anchor-based GRPO training process consists of five main steps, as summarized in Algorithm~\ref{alg:rl_pipeline}. Given multimodal inputs $(\mathbf{I}, \mathbf{P}, \mathbf{T})$, the pooled feature is perturbed to enable stochastic exploration:
    \begin{equation}
        \mathbf{f}_i = \phi_\text{attn\_pool}(\mathbf{I}, \mathbf{P}, \mathbf{T}) + \xi_i, \quad \xi_i \sim \mathcal{N}(0, \sigma^2)
    \end{equation}
where $\xi_i$ denotes Gaussian noise. 
The action head then generates candidate trajectories $\{\tau_i\}_{i=1}^N$:
    \begin{equation}
        \tau_i = \phi_\text{action}(\mathbf{f}_i), \quad i = 1, \dots, N
    \end{equation}

    Each trajectory $\tau_i$ is evaluated to obtain its reward $r_i = r(\tau_i)$, forming $\{r_i\}_{i=1}^N$, while the positive and negative anchors are defined as $r_\text{pos} = r(\tau_\text{pos})$ and $r_\text{neg} = r(\tau_\text{neg})$. The paired supervision signal $(\tau_{\text{neg}}, \tau_{\text{pos}})$ serves as a contrastive reference, encouraging the model to approach safe behaviors and avoid risky ones.

    If $\max(r_i)$ does not satisfy the safety threshold $\delta$, a feedback refinement process is triggered with the negative analysis block (detailed in Section~\ref{subsec:Negative-Enhanced SFT}), which is constructed from $\tau_{\text{pos}}$ and the worst-performing trajectory in $\{r_i\}_{i=1}^N$. This feedback produces corrected trajectories $\{\tau_j^\text{rec}\}_{j=1}^K$ with corresponding rewards $\{r_j^\text{rec}\}_{j=1}^K$. 
    This self-correction ability, learned in the negative-enhanced SFT, enables experience-guided trajectory improvement.
    Then, to calculate the group advantage, all trajectories are aggregated into a unified group:
    \begin{equation}
    \mathcal{T} = \{\tau_i\}_{i=1}^N \cup \{\tau_j^\text{rec}\}_{j=1}^K \cup \{\tau_{\text{pos}},\ \tau_{\text{neg}}\}
    \end{equation}
    
    The normalization statistics are computed exclusively from the policy-sampled rewards $R_{\text{sample}} = \{r_i\}_{i=1}^N \cup \{r_j^\text{rec}\}_{j=1}^K$, keeping the anchor rewards $r_{\text{pos}},\ r_{\text{neg}}$ out of the statistics to avoid distributional bias. Group-relative advantages are then assigned uniformly:
    \begin{equation}
        \hat{A}_k = \frac{r_k - \mathrm{mean}(R_{\text{sample}})}{\mathrm{std}(R_{\text{sample}})}, \quad \forall\, r_k \in R_{\text{sample}} \cup \{r_{\text{pos}},\ r_{\text{neg}}\}
    \end{equation}

    The policy $\theta = \{\phi_{\text{attn\_pool}},\ \phi_{\text{action}}\}$ is updated by minimizing an advantage-weighted dual-branch loss over all trajectories in $\mathcal{T}$:
    \begin{equation}
        \mathcal{L}
        =
        \frac{1}{|\mathcal{T}|}
        \sum_{\tau_k \in \mathcal{T}}
        \begin{cases}
        \hat{A}_k \cdot \mathcal{L}_{\mathrm{Huber}}(\tau_k, \tau_{\mathrm{pos}}),
        & \hat{A}_k \ge 0, \\
        |\hat{A}_k| \cdot \max(0, m - d(\tau_k,\tau_{\mathrm{neg}})),
        & \hat{A}_k < 0.
        \end{cases}
    \end{equation}
where $m$ is a margin distance used to keep generated trajectories sufficiently far from the negative anchor $\tau_{\mathrm{neg}}$, and $d(\cdot, \cdot)$ denotes the standard $L_2$ distance metric quantifying the spatial dissimilarity between two trajectory sequences. $\mathcal{L}_{\mathrm{Huber}}(\cdot, \cdot)$ denotes the Huber loss:
\begin{equation}
    \mathcal{L}_{\mathrm{Huber}}(\tau_k, \tau_{\mathrm{pos}}) = 
    \begin{cases} 
    \frac{1}{2} \left\| \tau_k - \tau_{\mathrm{pos}} \right\|_2^2, & \text{if } \left\| \tau_k - \tau_{\mathrm{pos}} \right\|_1 \le \delta_{\rm Huber}, \\
    \delta_{\rm Huber} \left( \left\| \tau_k - \tau_{\mathrm{pos}} \right\|_1 - \frac{1}{2} \delta_{\rm Huber} \right), & \text{otherwise},
    \end{cases}
\end{equation}
where $\delta_{\rm Huber}$ is a tunable threshold controlling the transition point.

\subsubsection{Reward Design}

Considering the multi-objective nature of autonomous driving, we define a weighted reward function that jointly accounts for trajectory quality, risk avoidance, and goal-directed efficiency:
\begin{equation}
r(\tau_i) = w_\text{traj} \cdot r_\text{traj}(\tau_i) + w_\text{pref} \cdot r_\text{pref}(\tau_i) + w_\text{goal} \cdot r_\text{goal}(\tau_i).
\end{equation}
where $r_\text{traj}(\tau_i)$ evaluates the quality of the trajectory based on PDMS (detailed in Section~\ref{subsec:generation}).

The preference reward term, $r_{\text{pref}}(\tau_i)$, serves as a relative safety score evaluated by the normalized spatial deviation from the negative baseline. Specifically, this reward explicitly encourages the predicted candidate trajectory $\tau_i$ to closely approximate the counterfactual positive anchor $\tau_{\text{pos}}$ while actively distancing itself from the high-risk negative anchor $\tau_{\text{neg}}$:
\begin{equation}
r_{\text{pref}}(\tau_i) =
\frac{1}{c_{\text{clip}}}
\operatorname{clip}\left(
\frac{d(\tau_i, \tau_{\text{neg}})}
{d(\tau_{\text{pos}}, \tau_{\text{neg}}) + \eta},
0,
c_{\text{clip}}
\right)
\end{equation}
where $\eta$ is a small constant introduced to ensure numerical stability, and
$c_{\text{clip}}$ is the upper bound for reward clipping.

Finally, the goal-directed reward $r_{\text{goal}}(\tau_i)$ promotes driving efficiency by penalizing the spatial deviation between the predicted trajectory's endpoint and the reference safe endpoint. This mechanism effectively prevents overly conservative behaviors while ensuring steady route progress. The reward is formulated based on the final displacement error (FDE) and decays exponentially:
\begin{equation}
    r_{\text{goal}}(\tau_i) = \exp\left(-\frac{\text{FDE}}{s_{\text{goal}}}\right)
\end{equation}
where $\text{FDE} = \|\tau_{i, T} - \tau_{\text{pos}, T}\|_2$. $\tau_{i, T}$ and $\tau_{\text{pos}, T}$ denote the terminal spatial coordinates at the final prediction step $T$ for the candidate trajectory $\tau_i$ and the positive anchor $\tau_{\text{pos}}$, respectively. The parameter $s_{\text{goal}}$ acts as a scaling factor that controls the sensitivity of the penalty, ensuring a smooth reward landscape for goal-reaching behaviors.

\section{Experiments}
\label{sec:experiments}

\subsection{Benchmarks}

\textbf{NAVSIM.}
NAVSIM \cite{dauner2024navsim} combines real-world sensor data with a simulator built on {OpenScene} \cite{peng2023openscene}. It provides multi-view camera images and fused point clouds, with approximately 103k training samples and 12k test samples. The benchmark mainly evaluates trajectory planning quality using a unified metric covering safety, progress, and comfort. Following the TTC criterion in Section~\ref{subsec:generation}, we further split the training set into positive and negative subsets, obtaining around 35k negative samples for negative-enhanced training.

\textbf{DeepAccident.}
DeepAccident \cite{wang2024deepaccident} is a CARLA-based benchmark for accident understanding and risk-aware decision-making, containing diverse collision scenarios with multi-view RGB cameras and LiDAR. The processed ego-centric subset includes 2,496 samples, with 948 collision cases. It evaluates language reasoning, risk prediction, trajectory error, and collision rate. In this work, it is used to assess semantic reasoning and risk recognition abilities beyond trajectory planning.

\subsection{Implementation Details}

\textbf{Dataset Splits.}
We follow the official splits of NAVSIM, using Navtrain for training and Navtest for evaluation. For DeepAccident, we use the public training and test subsets.

\textbf{Input and Output Settings.}
 We denote the history, short-term, and prediction horizons as $T_h=1.5$\,s, $T_s=1$\,s, and $T_p=4$\,s. The model takes 1.5\,s historical observations as input, including ego states and images from three forward-facing cameras (front, front-left, front-right) sampled at 2\,Hz. It predicts future ego trajectories over a 4\,s horizon at 2\,Hz.

\textbf{Model Architecture.}
We adopt {Qwen2.5-VL} as the backbone, apply LoRA for parameter-efficient fine-tuning, and attach a multi-layer perceptron (MLP)-based action head for trajectory generation. Please refer to \ref{sec:appendix1} for further model architecture details.

\textbf{Training Details.}
SafeAlign-VLA is trained with a progressive post-training workflow. For CSP-based risk labeling, the TTC threshold $TTC_{\text{risk}}$ is set to $2.0\,\mathrm{s}$ to identify negative samples, motivated by forward collision warning protocols\cite{ntsb2015forwardcas}. Stage 1 curriculum SFT is conducted for 15 and 5 epochs across its sequential substages, respectively. Stage 2 anchor-based GRPO is trained for 10 epochs. All experiments are conducted on four NVIDIA A6000 GPUs. Please refer to \ref{sec:appendix2} for further training details.

\subsection{Comparison with SOTA Methods}

Tables~\ref{tab:main_results_navsim} and \ref{tab:main_results_deepaccident} present the quantitative comparisons between the proposed method and SOTA baselines on the NAVSIM and DeepAccident benchmarks, respectively. Our method achieves a competitive performance on the NAVSIM benchmark, obtaining a PDMS of 89.1 using only camera inputs. Despite not using LiDAR, SafeAlign-VLA still surpasses several camera+LiDAR methods in overall planning quality, demonstrating strong trajectory planning capability in balancing safety, progress, and comfort. 

On the DeepAccident benchmark, SafeAlign-VLA consistently outperforms competitive VLM baselines across language reasoning, risk prediction, and trajectory planning metrics. In particular, it achieves 84.2\% language accuracy and 85.8\% risk prediction accuracy, indicating superior scene understanding and hazard recognition ability. Meanwhile, it attains the lowest trajectory errors and collision rate, reducing the collision rate to 3.36\%, which demonstrates effective proactive decision-making in safety-critical scenarios.

\begin{table}[t!]
\centering
\caption{Performance Comparison of Trajectory Prediction Models}
\label{tab:main_results_navsim}
\begin{threeparttable}
\footnotesize
\begin{tabular}{l|c|cc|ccccc|c}
\toprule
\textbf{Method} & \textbf{Ref} & \multicolumn{2}{c|}{\textbf{Input}} & \textbf{NC} & \textbf{DAC} & \textbf{EP} & \textbf{TTC} & \textbf{C} & \textbf{PDMS} \\
\cmidrule(r){3-4}
 & & Camera & LiDAR & & & & & & \\
\midrule
\multicolumn{10}{l}{\textit{E2E Methods}} \\
\midrule
VADv2 \cite{chen2024vadv2} &  ICLR'26    & \checkmark &            & 97.9 & 91.7 & 77.6 & 92.9 & \textbf{100}  & 83.0 \\
UniAD \cite{hu2023planning} & CVPR'23  & \checkmark &            & 97.8 & 91.9 & 78.8 & 92.9 & \textbf{100}  & 83.4 \\
Transfuser \cite{prakash2021multi} & TPAMI'23  & \checkmark & \checkmark & 97.7 & 92.8 & 79.2 & 92.8 & \textbf{100}  & 84.0 \\
PARA-Drive \cite{10656117} & CVPR'24  & \checkmark &            & 97.9 & 92.4 & 79.3 & 93.0 & 99.8 & 84.0 \\
ARTEMIS  \cite{feng2025artemis} & IEEE'25 & \checkmark & \checkmark & 98.3 & 95.1 & 81.4 & 94.3 & \textbf{100}  & 87.0 \\
DiffusionDrive \cite{liao2025diffusiondrive} & CVPR'25 & \checkmark & \checkmark & 98.2 & 96.2 & \textbf{82.2} & 94.7 & \textbf{100}  & 88.1 \\
WoTE \cite{li2025end} & ICCV'25 & \checkmark & \checkmark & 98.5 & 96.8 & 81.9 & 94.9 & 99.9 & 88.3 \\
\midrule
\multicolumn{10}{l}{\textit{VLM Methods}} \\
\midrule
InternVL3-2B$^\dagger$ \cite{zhu2025internvl3} & arXiv'25 & \checkmark &            & 97.2 & 87.4 & 78.9 & 91.8 & \textbf{100} & 78.9 \\
QwenVL2.5-3B$^\dagger$ \cite{bai2025qwen3} & arXiv'25  & \checkmark &            & 97.7 & 93.1 & 73.0 & 92.3 & \textbf{100} & 84.2 \\
\midrule
\multicolumn{10}{l}{\textit{VLA Methods}} \\
\midrule
DrivingGPT \cite{chen2025drivinggpt}  & ICCV'25   & \checkmark &            & \textbf{98.9}   & 90.7   & 79.7   & 94.9   & 95.6   & 82.4   \\
RoboTron-Sim \cite{11444934}  & ICCV'25  & \checkmark &      & 98.2 & 93.6 & 81.1 & 93.8 & 99.9 & 85.6  \\
AutoVLA  \cite{zhou2026autovla}  & NeurIPS'25 & \checkmark &            & 98.4   & 95.6   & 81.9   & 98.0   & 99.9   & \textbf{89.1}  \\
\midrule
\textbf{SafeAlign-VLA (Ours)}  &   & \checkmark &  & 98.6   & \textbf{97.2} & 81.7  & \textbf{98.1}  & \textbf{100}  & \textbf{89.1} \\
\bottomrule
\end{tabular}%
\begin{tablenotes}
  \footnotesize
  \item[$\dagger$] Fine-tuned on NAVSIM navtrain split using the respective pre-trained VLM backbone.
\end{tablenotes}
\end{threeparttable}
\end{table}

\FloatBarrier

\begin{table}[t!]
\centering
\caption{Quantitative comparison on the DeepAccident validation set.}
\label{tab:main_results_deepaccident}
\footnotesize
\begin{tabular}{lcccccc}
\toprule
& \multicolumn{1}{c}{Language} 
& \multicolumn{2}{c}{Risk Prediction} 
& \multicolumn{3}{c}{Output Trajectory} \\
\cmidrule(lr){2-2}
\cmidrule(lr){3-4}
\cmidrule(lr){5-7}
Method 
& Acc.  
& Acc.  
& Rec. 
& L2@1s 
& L2@4s  
& Coll.  \\
& (\%)$\uparrow$ 
& (\%)$\uparrow$ 
& (\%)$\uparrow$ 
& (m)$\downarrow$ 
& (m)$\downarrow$ 
& (\%)$\downarrow$ \\
\midrule
LLaVA-1.5 (7B)        & 63.5 & 70.2 & 58.1 & 0.56 & 3.04 & 7.29 \\
Llama-3.2-V (11B)     & 69.8 & 75.6 & 63.7 & 0.45 & 2.40 & 5.91 \\
DeepSeek-VL (7B)      & 75.1 & 79.8 & 69.2 & 0.48 & 2.51 & 6.10 \\
InternVL-2.5 (8B)     & 72.4 & 76.9 & 63.5 & 0.44 & 2.35 & 5.85 \\
Qwen2.5-VL (7B)       & 75.6 & 82.3 & 64.7 & 0.40 & 2.30 & 5.13 \\
\midrule
\textbf{SafeAlign-VLA (Ours)}         & \textbf{84.2} & \textbf{85.8} & \textbf{81.9} 
& \textbf{0.32} & \textbf{1.78} & \textbf{3.36} \\
\bottomrule
\end{tabular}%
\end{table}

\subsection{Ablation Studies}
\label{subsec:ablation}

To validate the effectiveness of each component, we conduct progressive ablation studies by gradually introducing modules based on the previous setting.
Specifically, {ID-1} is the baseline model trained with standard SFT (Base SFT) on raw data. {ID-2} introduces the {CSP} module into SFT, which separates positive and negative samples to achieve the negative-enhanced curriculum SFT. {ID-3} further adds a standard GRPO-based RL. Based on ID-3, {ID-4} adds the feedback refinement module while {ID-5} introduces the anchor-based reward guidance using positive and negative trajectories as corresponding anchors. {ID-6} is the full model combining both feedback refinement and reward anchors.

We design two groups of ablations. The {first group} (ID-1, ID-2, ID-3, ID-6) evaluates the gains from negative-enhanced SFT and GRPO optimization. The {second group} (ID-3, ID-4, ID-5, ID-6) analyzes the contributions of {feedback refinement} and {reward anchors} within GRPO.
Table~\ref{tab:ablation_main} presents the ablation results of SafeAlign-VLA. The CSP module (ID-2) improves PDMS from 87.8 to 88.3 and achieves the best NC and TTC scores, showing that structured safety semantics and counterfactual positive supervision effectively enhance safety-oriented planning while avoiding unsafe imitation from raw data. DAC and EP slightly decrease, as this stage prioritizes safety. Adding RL (ID-3) further improves EP, DAC, and PDMS, indicating that the model benefits from transitioning from passive imitation to active policy exploration. The full model (ID-6) achieves the best overall performance, validating the effectiveness of integrating feedback refinement and anchor-based preference learning into GRPO.

Table~\ref{tab:ablation_feedback_anchor} further analyzes the GRPO components. ID-4 brings moderate gains over ID-3, while ID-5 achieves larger improvements, especially on TTC and PDMS, indicating that anchor-based preference signals provide stronger optimization guidance. Combining both strategies in ID-6 yields the best performance, demonstrating their complementary benefits.
\begin{table}[t!]
\centering
\caption{Progressive ablation results of the proposed framework. Higher is better for all metrics.}
\label{tab:ablation_main}
\footnotesize
\begin{tabular}{l|l|ccccc|c}
\toprule
\textbf{ID} & \textbf{Method} & \textbf{NC $\uparrow$} & \textbf{DAC $\uparrow$} & \textbf{EP $\uparrow$} & \textbf{TTC $\uparrow$} & \textbf{C $\uparrow$} & \textbf{PDMS $\uparrow$} \\
\midrule
ID-1 & Base SFT & 98.2 & 97.0 & 81.1 & 95.2 & 99.9 & 87.8 \\
ID-2 & + CSP & \textbf{98.8} & 96.8 & 80.4 & \textbf{98.4} & 99.8 & 88.3 \\
ID-3 & + RL Optimization & 98.4 & \textbf{97.2} & 81.6 & 96.1 & \textbf{100.0} & 88.5 \\
\midrule
ID-6 &\textbf{+ Full GRPO (SafeAlign-VLA)}& 98.6 & \textbf{97.2} & \textbf{81.7} & 98.1 & \textbf{100.0} & \textbf{89.1} \\
\bottomrule
\end{tabular}
\end{table}

\FloatBarrier

\begin{table}[t!]
\centering
\caption{Ablation study on the effects of feedback refinement and reward anchors in GRPO.}
\label{tab:ablation_feedback_anchor}
\footnotesize
\begin{tabular}{l|cc|ccccc|c}
\toprule
\textbf{ID} & \textbf{Feedback} & \textbf{Anchor} & \textbf{NC $\uparrow$} & \textbf{DAC $\uparrow$} & \textbf{EP $\uparrow$} & \textbf{TTC $\uparrow$} & \textbf{C $\uparrow$} & \textbf{PDMS $\uparrow$} \\
\midrule
ID-3 &  &  & 98.4 & 97.2 & 81.6 & 96.1 & \textbf{100.0} & 88.5 \\
ID-4 & \checkmark &  & 98.4 & \textbf{97.3} & 81.4 & 96.5 & \textbf{100.0} & 88.6 \\
ID-5 &  & \checkmark & 98.5 & 97.2 & 81.6 & 97.8 & \textbf{100.0} & 88.9 \\
\midrule
ID-6 & \checkmark & \checkmark & \textbf{98.6} & 97.2 & \textbf{81.7} & \textbf{98.1} & \textbf{100.0} & \textbf{89.1} \\
\bottomrule
\end{tabular}
\end{table}

\subsection{Qualitative Analysis}
\label{subsec:qualitative}

\begin{figure*}[t!]
    \centering
    \begin{subfigure}{0.89\textwidth}
        \centering
        \includegraphics[width=\linewidth]{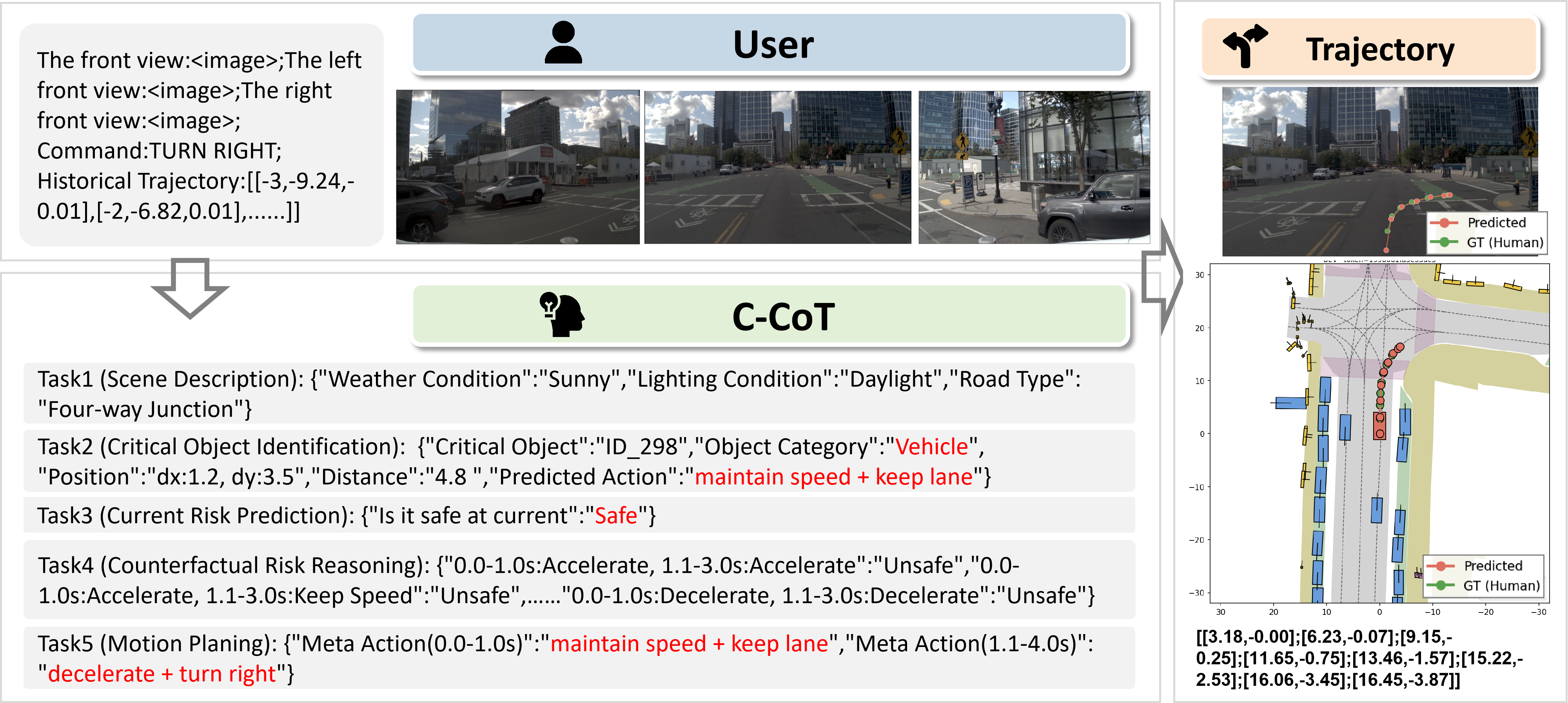}
        \caption{Right-turn case: The model assesses the scene as safe and maintains initial speed before turning.}
        \label{fig:qualitative_a}
    \end{subfigure}
    \hfill
    \begin{subfigure}{0.89\textwidth}
        \centering
        \includegraphics[width=\linewidth]{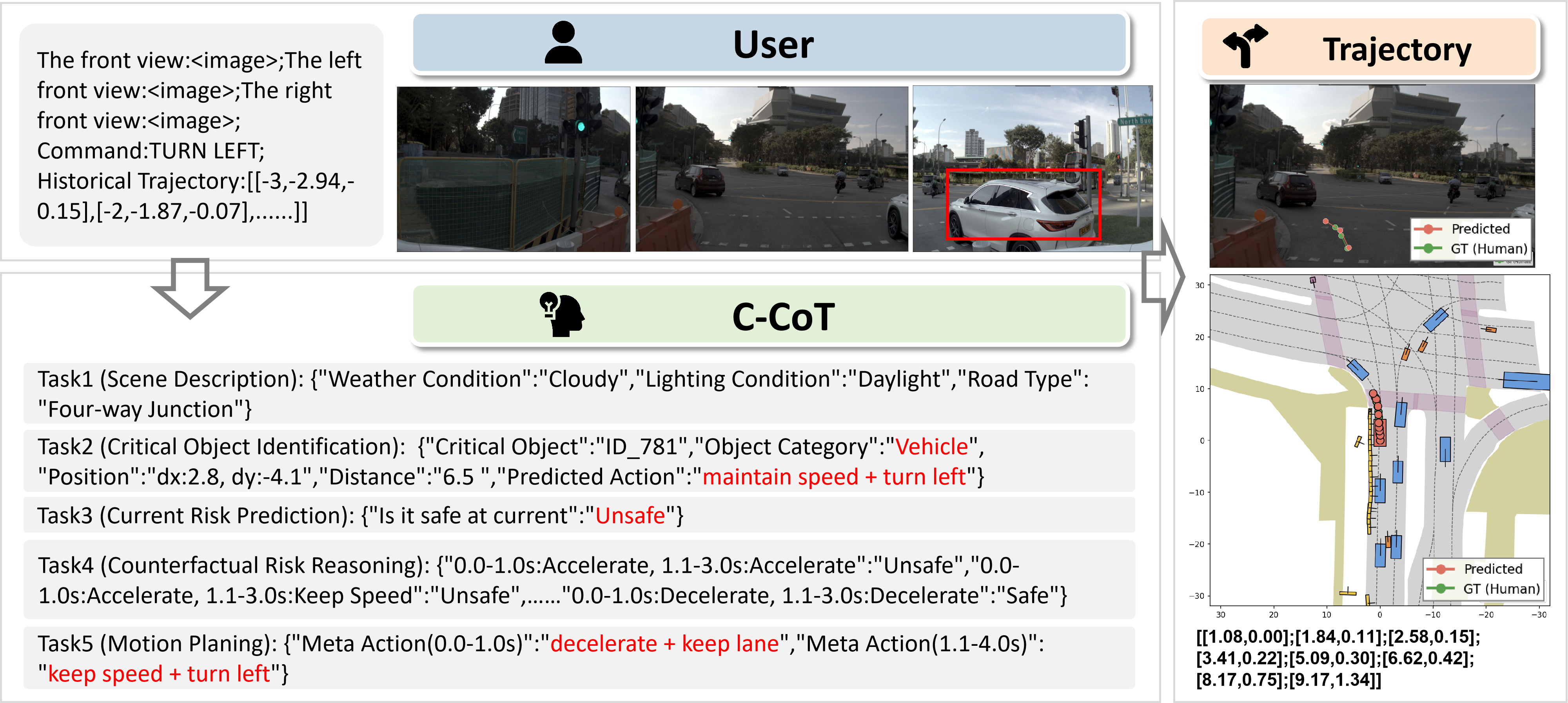}
        \caption{Left-turn case: The model identifies a crossing vehicle as a risk and adopts a yield-first strategy.}
        \label{fig:qualitative_b}
    \end{subfigure}

    \caption{Qualitative results of our framework in urban intersection scenarios. (a) A safe right-turn scenario where the model aligns with human reference trajectories. (b) A complex left-turn scenario requiring counterfactual reasoning to yield to a critical object. Both cases demonstrate interpretable and human-like planning.}
    \label{fig:qualitative}
\end{figure*}

Figure~\ref{fig:qualitative} shows two representative intersection cases, including model inputs, counterfactual CoT reasoning outputs, and final predicted trajectories.
In Figure~\ref{fig:qualitative_a}, the model judges the current situation as safe. It therefore maintains speed initially and then decelerates to complete the right turn, generating a trajectory close to the human reference.
In Figure~\ref{fig:qualitative_b}, the crossing vehicle is identified as the critical object, leading to an unsafe risk assessment. After counterfactual reasoning, the model adopts a conservative strategy by decelerating first to yield, then completing the left turn after the conflict is resolved. These examples demonstrate that our framework can identify critical risk sources, perform counterfactual safety reasoning, and generate interpretable, human-like decisions in complex urban intersections.

\begin{figure}[t!]
\centering
\includegraphics[width=\textwidth]{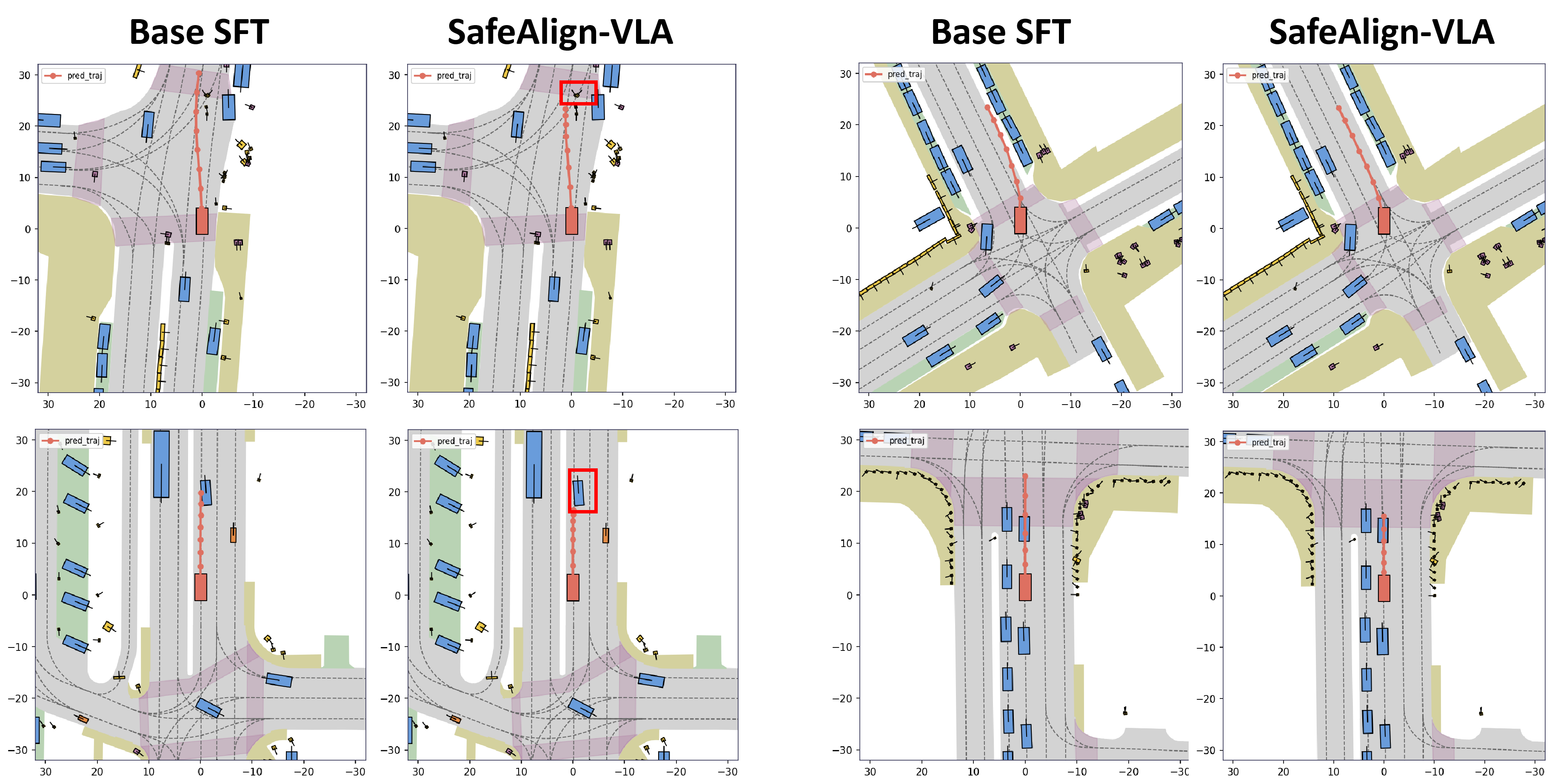}
\caption{Qualitative comparison of predicted trajectories between Base SFT (ID-1) and Full GRPO (ID-6) in some representative conflict scenarios.}
\label{fig:ablation_case_study}
\end{figure}

The ablation visualization (Figure~\ref{fig:ablation_case_study}) further demonstrates the benefits of our approach. Compared to the baseline SFT model, our framework proactively detects critical risk sources, such as crossing vehicles and potential intersection conflicts, and adjusts trajectories in advance to avoid collisions. The predicted paths are smoother and more continuous, with reduced lateral oscillations and improved longitudinal control. Trajectories also exhibit better road-fitting, remaining within lane boundaries, and avoiding off-road encroachment. Overall, the model enhances safety, respects driving constraints, and generates human-like, executable trajectories in complex urban scenarios.

\section{Conclusion}
\label{sec:conclusion}

In this paper, we propose SafeAlign-VLA, a unified negative-enhanced safe alignment framework designed to enhance the safety, interpretability, and robustness of VLA models in autonomous driving. Recognizing that existing end-to-end models predominantly rely on positive expert demonstrations and struggle to generalize in safety-critical, long-tail scenarios, our approach explicitly integrates structured negative supervision into both the SFT and RL finetune paradigms. We developed a counterfactual safety pairing data construction paradigm driven by C-CoT reasoning to autonomously extract failure modes and generate corresponding safe alternatives. Building upon this, we introduced a negative-enhanced curriculum  SFT process for precise corrected analysis and trajectory correction, followed by an anchor-based GRPO post-training. This RL strategy distinctively leverages both positive and negative trajectories as contrastive anchors, effectively penalizing high-risk behaviors and steering the policy toward optimal safety.  

Extensive evaluations on the NAVSIM and DeepAccident benchmarks thoroughly validate the efficacy of our framework. SafeAlign-VLA establishes a new SOTA on the NAVSIM testset with a PDMS of 89.1, highlighting its superior capacity to balance safety, driving progress, and passenger comfort. Furthermore, on the safety-critical DeepAccident benchmark, our method significantly outperformed competitive VLM baselines, reducing the collision rate to a remarkable 3.36\% while achieving exceptional hazard recognition (85.8\%) and language reasoning (84.2\%) accuracies. Ablation studies confirmed that the synergistic combination of counterfactual negative supervision and anchor-based reward guidance is crucial for successfully transitioning policies from passive imitation to proactive safety optimization.  

Ultimately, SafeAlign-VLA demonstrates that explicitly incorporating real-world negative signals, rather than merely imitating positive demonstrations, is essential for establishing clear safety boundaries and ensuring reliable decision-making in complex environments. 
Future work could integrate generative world models to simulate future environmental dynamics explicitly. This paradigm shift would elevate the negative-enhanced learning pipeline from the semantic and trajectory levels to the pixel level, further bridging the gap between high-level semantic reasoning and precise physical execution in autonomous driving systems. 

\section*{Acknowledgement(s)}
The research is supported by the National Natural Science Foundation of China (No. 52325209, 52272420, 52532010) and the Tsinghua-Toyota Joint Research Institute Inter-disciplinary Program.

\newpage
\bibliographystyle{elsarticle-num}
\bibliography{ref}

\newpage
\appendix
\section{Implementation Details}
\label{app:implementation}

\subsection{Network Architecture}\label{sec:appendix1}

Our model architecture consists of the following components:

\begin{itemize}
    \item {Vision encoder}: \textit{Qwen2.5-VL-7B} serves as the backbone, extracting visual features from the multi-camera inputs.
    \item {Action head}: A three-layer MLP projects the pooled VLM feature to 256 dimensions and predicts future waypoints over the 4-second horizon.
    \item {Parameter-efficient fine-tuning}: LoRA with $r=64$ is applied to the language backbone for efficient adaptation in SFT, covering both the self-attention and feed-forward sublayers.
\end{itemize}

\subsection{Training Details}\label{sec:appendix2}

All multi-view camera images are resized to $224 \times 224$ pixels before being fed into the vision encoder. The maximum sequence length for the text tokenizer is $4096$ tokens to accommodate the extensive C-CoT reasoning sequences. Training is conducted in two stages:

\paragraph{Stage 1: Supervised Fine-Tuning}
\begin{itemize}
    \item Optimizer: AdamW
    \item Learning rate: $2\times10^{-5}$ for LoRA parameters, $1\times10^{-4}$ for the action head
    \item Curriculum SFT: 15 epochs for the basic SFT substage and 5 epochs for the negative-enhanced SFT substage on four NVIDIA A6000 GPUs 
\end{itemize}

\paragraph{Stage 2: RL Fine-Tuning}
\begin{itemize}
    \item Trajectory group size: $(6+2)$, including six sampled candidates, and two positive/negative anchors
    \item Feedback threshold: $\delta=0.6$
    \item RL hyperparameters: $m=1.0$, $\eta=10^{-6}$, $c_{\text{clip}}=1.5$, $s_{\text{goal}}=3.0$
    \item Reward weights: $w_{traj} = 0.5$, $w_{pref} = 0.3$, $w_{goal} = 0.2$
    \item Post-training: 10 epochs on four NVIDIA A6000 GPUs
\end{itemize}

\end{document}